%% file: main_new.tex
\documentclass{article}
\usepackage{graphicx} 
\usepackage{enumerate}
\usepackage{amsmath}
\usepackage{amssymb}
\usepackage[a4paper, margin=1in]{geometry} 
\usepackage{comment}
\usepackage{algorithm}
\usepackage{algpseudocode}
\usepackage[utf8]{inputenc}
\usepackage{natbib}
\usepackage{xcolor}
\usepackage{xparse} 
\usepackage{amsfonts} 
\usepackage{enumitem}
\usepackage{enumerate}
\input{macro}

\title{Hierarchical Reinforcement Learning for Optimal Agent Grouping in Cooperative Systems}
\author{Liyuan Hu \\ Department of Statistics, London School of Economics\\ l.hu11@lse.ac.uk}
\date{}
\begin{document}

\maketitle

\begin{abstract}
This paper presents a hierarchical reinforcement learning (RL) approach to address the agent grouping or pairing problem in cooperative multi-agent systems. The goal is to simultaneously learn the optimal grouping and agent policy. By employing a hierarchical RL framework, we distinguish between high-level decisions of grouping and low-level agents' actions. Our approach utilizes the CTDE (Centralized Training with Decentralized Execution) paradigm, ensuring efficient learning and scalable execution. We incorporate permutation-invariant neural networks to handle the homogeneity and cooperation among agents, enabling effective coordination. The option-critic algorithm is adapted to manage the hierarchical decision-making process, allowing for dynamic and optimal policy adjustments. 
\end{abstract}

\section{Introduction}

Sequential grouping or pairing problems are ubiquitous in many domains, from resource allocation and scheduling to team formation and matchmaking. These problems are challenging due to their inherent combinatorial complexity and the need for efficient decision-making under constraints.

 In studies of natural \citep{jeanson2005emergence,wittemyer2007hierarchical} and multi-agent systems \citep{phan2021vast}, grouping has been recognized as a means to enhance efficient collaboration. However, devising a general criterion for grouping agents without specific domain knowledge remains a significant challenge for the research community. Previous works often focus on well-structured tasks with predefined roles or task decompositions \citep{cossentino2014handbook, lhaksmana2018role, phan2021vast,jiang2021multi}, which may not be practical in real-world settings and can restrict the transferability of these methods. 
Existing approaches to grouping and pairing policies in RL have limitations. For example, user pairing in the NOMO system \citep{Wang2021JointRM, s20247094} suffers from exponential growth in the state-action space when modeling Q(s, a) directly with actions as inputs. Other RL-based methods \citep{10.1016/j.eswa.2016.07.047} focus on grouping problems but do not address sequential decision-making. Grouping problems typically involve partitioning a set into mutually disjoint subsets according to specific criteria and constraints, yet these approaches lack scalability and flexibility for sequential applications.
Recently, \citet{saul2005advances} proposed an automatic agent clustering method based on the contribution weight of each agent to specific groups. 
Our method differ with their work in a sense that the policy of agent grouping is learnt under RL framework with the goal of maximizing the cumulative reward while they learn the grouping by a more indirect sparsity regulation.

The application of RL to multi-agent matching or pairing problems presents unique computational and modeling challenges, particularly when scaling to larger problem sizes. Traditional Q-learning or option-critic approaches face critical limitations in this context:

\begin{enumerate}
    \item \textbf{Challenges in Critic Network Design and Learning:}
    \begin{itemize}
        \item \textbf{Expansive Option Space:} The number of ways to pair $n$ items (teams) is given by the double factorial $(n-1)!!$, leading to a massive growth in the option space. For example, when $n=20$, the number of pairings exceeds 3.7 billion.
        \item \textbf{Inefficient Modeling:} Pairings such as $(1,2)$ and $(3,4)$ versus $(3,4)$ and $(1,2)$ represent identical team matchings but may be treated differently in conventional Q-function models, leading to redundancy and inefficiency.
    \end{itemize}

    \item \textbf{Challenges in Policy Network Design and Learning:}
    \begin{itemize}
        \item \textbf{Expansive Action Space:} As the number of teams and team sizes increase, the action space grows exponentially, complicating policy learning.
        \item \textbf{Scalability Issues:} Traditional option-critic architectures struggle to scale due to the exponential growth in policy parameter space with increasing option complexity.
    \end{itemize}
\end{enumerate}

To address these challenges, we propose the following contributions:

\begin{enumerate}
    \item \textbf{Solutions to Challenges in Critic Network Design:}
    \begin{itemize}
        \item We redefine the critic Q-function by embedding pairing information directly within the network architecture rather than treating it as a separate input. This ensures permutation invariance of team pairings, enhancing modeling efficiency and reducing computational redundancy (see Figure \ref{fig:PI critic}).
        \item We decompose the joint Q-function into combinations of pair Q-values, transforming the problem of finding optimal team pairings into a linear optimization problem. This significantly reduces computational complexity by avoiding exhaustive enumeration (see Section \ref{sec:network arch}).
    \end{itemize}

    \item \textbf{Solutions to Challenges in Policy Network Design:}
    \begin{itemize}
        \item We implement a Permutation Invariant (PI) network architecture that leverages the homogeneity of agents for effective parameter sharing (see Figure \ref{fig:PI network}).
        \item Instead of using independent policy networks for each option, we incorporate pairing information directly within the policy network architecture (see Figure \ref{fig:PI network}).
    \end{itemize}
\end{enumerate}

Our critic and policy networks employ a shared-parameter embedding for individual states, which are aggregated to form team character embeddings. These embeddings are transformed to generate PI matching embeddings for team pairs, which are summed to produce a final Q-value through a single fully connected layer. This architecture avoids the curse of dimensionality by circumventing direct engagement with the expansive action space, ensuring scalability as the number of teams or individuals increases without a proportional rise in parameter count.

\section{Related Work}

\subsection{Hierarchical Reinforcement Learning (HRL)}
Hierarchical RL addresses the challenge of large state and action spaces in long-horizon tasks by decomposing the problem into a hierarchy of subtasks. A higher-level policy selects optimal subtasks, while lower-level policies solve them, improving exploration and scalability. 

\textbf{HRL Foundations.} \citet{pateria2021hierarchical} highlight how HRL improves exploration through structured subtasks or subgoals. The Semi-Markov Decision Process (SMDP) framework \citep{SUTTON1999181} underpins many HRL methods by allowing variable-length subtasks. The Option-Critic architecture \citep{Bacon2016TheOA, Chunduru2022AttentionO} automates the learning of options (subtasks) and their termination conditions, facilitating adaptive and autonomous decision-making. Extensions to multi-agent systems \citep{Chakravorty2019OptioncriticIC} allow decentralized execution and cooperative behavior. 

\textbf{Policy over Options.} An option is defined as a tuple $(\mathcal{I}, \pi, \beta)$, where $\mathcal{I}$ specifies the initiation set, $\pi$ is the intra-option policy, and $\beta$ is the termination function. Our work builds on this foundation by considering recursive optimality, where sub-behaviors are individually optimal, and policy learning over large option spaces, where permutation invariance and parameter efficiency are key.

\subsection{Multi-Agent Reinforcement Learning (MARL)}
Multi-agent RL involves multiple interacting agents learning policies to optimize collective or individual objectives. Hierarchical approaches \citep{pateria2021hierarchical} integrate HRL concepts to decompose complex team-based decision problems, enabling more efficient coordination and scalability. 

\subsection{Matching Problems}
Matching problems, particularly in reinforcement learning, span static and dynamic contexts, bipartite and monopartite structures, and stochastic settings. 

\textbf{Bipartite vs. Monopartite Matching.} Bipartite matching involves two distinct sets (e.g., tasks and workers), while monopartite matching operates within a single homogeneous set. Dynamic bipartite graph matching (DBGM) \citep{wang2019adaptive} addresses scenarios where tasks and workers arrive and leave dynamically. The proposed adaptive batch-based framework leverages RL for batch splitting, enabling theoretical guarantees and efficient real-time decision-making.

\textbf{Order Dispatch and Markov Matching Markets.} Order dispatch systems \citep{xu2018large} apply policy iteration and Q-learning to optimize driver-task assignments, handling stochastic action spaces \citep{cohen2022dynamic}. Markov matching markets \citep{min2022learn} introduce stochastic agent sets and planner actions that affect environmental transitions, integrating RL to optimize matching policies.

\textbf{Competing Matching with Complementary Preferences.} \citet{huang2022two} tackle many-to-one matching using Thompson Sampling for preference exploration and double matching for stability, demonstrating RL's applicability in balancing exploration and exploitation.

\subsection{The Option-Critic Architecture}
The Option-Critic framework \citep{Bacon2016TheOA} provides a foundation for hierarchical policy learning by automating the discovery of options and their termination conditions. However, traditional architectures face challenges in scaling to large option spaces.

\textbf{Differences in Proposed Architecture.}
\begin{enumerate}
    \item \textbf{Option Space:} Our problem involves an extremely large, known option space, necessitating parameter-efficient critic networks and advanced optimization methods to avoid exhaustive enumeration.
    \item \textbf{Termination Function:} Unlike traditional approaches requiring learned termination functions, our framework assumes predefined termination conditions, simplifying the learning process.
\end{enumerate}

\textbf{Comparison to Multi-Agent Cooperative Option-Critic.} \citet{Chakravorty2019OptioncriticIC} address partially observed state spaces and decentralized execution in multi-agent settings. Our work contrasts by leveraging fully observed state spaces, homogeneity assumptions, and parameter-efficient critic designs to manage large option spaces effectively.

In summary, our approach integrates and extends hierarchical RL, multi-agent RL, and matching theory to address challenges in large-scale, dynamic decision-making tasks. By leveraging permutation-invariant architectures and efficient policy optimization, we provide scalable solutions for complex pairing and grouping problems.

\section{Hierarchical Multi-Agent Reinforcement Learning for Agent Grouping/Pairing}

We first introduce the motivating example.
In the Intern Health Study \citep{wang2023effectiveness}, medical interns from diverse backgrounds—different specialties and universities—are weekly grouped into teams to compete on health metrics including daily step counts, mood scores, and sleep hours, all managed through a dedicated app. This app is responsible for determining the optimal team pairings each week and deciding whether to send health-promoting messages to individuals daily, with the goal of maximizing interns' health throughout their internship period. This setup presents a challenging multi-agent problem, where high-level decisions on team formations and low-level decisions on individual interventions must be dynamically optimized. The hierarchical nature of these decisions—where team pairings influence daily interactions and messaging—motivates the application of a hierarchical multi-agent reinforcement learning approach, capable of efficiently coordinating both levels to achieve the best health outcomes.

This problem is addressed within a hierarchical Reinforcement Learning (RL) framework, involving homogeneous and fully cooperative agents. This approach is well-suited for scenarios where agents need to dynamically form groups or pairs to optimize collective performance.

\begin{itemize}
    \item \textbf{Homogeneous and Fully Cooperative Agents}.
In this setting, agents are homogeneous, meaning they share the same structure, skills, and learning models. This homogeneity allows for efficient learning and scalability, as agents can share experiences and model parameters. Despite sharing the same policy network, agents can exhibit diverse behaviors due to their unique observations at any given time. Additionally, agents can leverage centralized value functions, incorporating mutual information and joint actions, to enhance learning efficiency \citep{Gronauer2021MultiagentDR}. We adopt the CTDE (Centralized Training with Decentralized Execution) paradigm, which is state-of-the-art for multi-agent learning \citep{kraemer2016multi,oliehoek2008optimal}.

\item \textbf{Hierarchical Reinforcement Learning} 
The hierarchical RL framework involves two levels of decision-making:
\begin{enumerate}
    \item \textbf{Option Policy}: This higher-level policy determines the groupings or pairings of agents (e.g., weekly team pairings), effectively selecting "options" that define the structure of interactions.
    \item \textbf{Intra-Option Policy}: Within the context of a selected grouping (option), this lower-level policy governs the day-to-day interactions (e.g., daily messaging) aimed at optimizing health outcomes.
\end{enumerate}

Given the availability of a centralized controller in our application (e.g., an IHS APP), option decisions can be based on the full state space, unlike scenarios where options are based on partial observations \citep{Chakravorty2019OptioncriticIC}. The homogeneity of agents suggests the use of permutation-invariant neural networks to reduce the parameter space of the joint Q-function or policy network.

\end{itemize}

\subsection{The Options Framework}
Define the state for subject $j$ of team $i$ observed at time $t$ as $\sijt$, the action as $\aijt$ and the reward as $\rijt$. We consider a discrete and finite individual action space $\mathcal{A} = \{a_1, \dots, a_{|\mathcal{A}|}\}$. Define the number of teams as $M$, the number of days requiring individual action decision each week as $H$ and the weeks are indicated by $w$.
We model the underlying system as an MDP and employ the options framework to structure hierarchical learning. 
An option $\omega$ is defined as a triple $(\mathcal{I}_\omega, \pi_\omega, \beta)$:
\begin{itemize}
\item Initiation Set $\mathcal{I}_\omega$: Specifies the states where option $\omega$ can be initiated.
    \item Intra-Option Policy

  $$
  \pi_\omega(\ba_t \mid \bs_t; \theta) = \prod_{k=1}^{M/2} \prod_{i \in m_k(\omega)} \prod_{j} \pi(\aijt \mid \bs_{m_k(\omega), t}; \theta)
  $$
  where $\bs_{m_k(\omega)} = \{\sijt[i][j][t]\}_{i \in m_k(\omega), j}$. 
  Define the set of team matching pairs induced by $\omega$ as $\mathcal{S}_{\omega}=\{(m_1(\omega)^1, m_1(\omega)^2), (m_2(\omega)^1, m_2(\omega)^2),\dots,(m_{M/2}(\omega)^1, m_{M/2}(\omega)^2): \text{where }\cup(m_i(\omega)^1, m_i(\omega)^2) = [M] \text{ and } (m_i(\omega)^1, m_i(\omega)^2) \cap (m_j(\omega)^1, m_j(\omega)^2)=\emptyset, \forall i\neq j\}$. Define $m_i(\omega)=(m_i(\omega)^1, m_i(\omega)^2)$.
  The parameter $\theta$ is shared across all teams and options to ensure consistency and reduce complexity.
\item Termination Condition $\beta$: Determines the probability of terminating the option in a given state:
  $$
  \beta(\Sijt[\cdot][\cdot][t]) = 
  \begin{cases} 
   0 & \text{if } wH \leq t < (w+1)H \\
   1 & \text{otherwise}
  \end{cases}
  $$
\end{itemize}

Agents select an option $\omega$ according to the policy over options $\pi_{\Omega}(\omega \mid s)$, execute the intra-option policy $\pi_\omega$ until termination (as dictated by $\beta$), and then repeat the process. This setup transforms the MDP into a Semi-Markov Decision Process (SMDP), enabling the use of optimal value functions over options $V_\Omega(s)$ and $Q_\Omega(s, \omega)$ 

We adopt the option-critic frame work to learn the joint Q network and intra-option policy. 
We define several important functions in the option-critic framework under our agent grouping settings.
The option-value function $Q_{\Omega, \psi}(s, \omega)$ is defined as:
$$
Q_{\Omega, \psi}(s, \omega) = \sum_{\ba} \pi_{\omega, \theta}(\ba \mid \bs) Q_{U, \psi}(\bs, \omega, \ba)
$$
where $Q_U$ represents the value of executing an action in the context of a state-option pair:
$$
Q_{U, \psi}(\bs, \omega, a) = \sum_i r(s_i, a_i) + \gamma \sum_{\bs'} P(\bs' \mid \bs, \ba, \omega) U(\omega, \bs')
$$
The option-value function upon arrival $U_\psi(\omega, \bs')$ is:
$$
U_\psi(\omega, \bs') = (1 - \beta(\bs')) Q_{\Omega, \psi}(\bs', \omega) + \beta(\bs') V_{\Omega, \psi}(\bs')
$$
with $V_{\Omega, \psi}(\bs) = \sum_{\omega} \pi_{\Omega}(\omega) Q_{\Omega, \psi}(\bs, \omega)$.

The gradient of the expected discounted return with respect to $\theta$ and initial condition $(\bs_0, \omega_0)$ is:
$$
\sum_{\bs, \omega} \mu_{\Omega}(\bs, \omega \mid \bs_0, \omega_0) \sum_{\ba} \frac{\partial \pi_{\omega, \theta}(\ba \mid \bs)}{\partial \theta} Q_U(\bs, \omega, \ba)
$$
where $\mu_{\Omega}(\bs, \omega \mid \bs_0, \omega_0)$ is the discounted weighting of state-option pairs along trajectories starting from $(\bs_0, \omega_0)$.

The gradient of the expected discounted return with respect to $\theta$ and initial condition $(\bs_0, \omega_0)$ is:
$$
\sum_{\bs, \omega} \mu_{\Omega}(\bs, \omega \mid \bs_0, \omega_0) \sum_{\ba} \frac{\partial \pi_{\omega, \theta}(\ba \mid \bs)}{\partial \theta} Q_U(\bs, \omega, \ba)
$$
where $\mu_{\Omega}$ 
$(\bs, \omega |$  
$\bs_0$, 
$\omega_0)$ is the discounted weighting of state-option pairs along trajectories starting from
 $(\bs_0, \omega_0)$.

With those definitions, the option-critic framework in \cite{Bacon2016TheOA} can be adopted to solve the hierarchical policy learning in theory. However, the large option/action space involved in the Q functions above make it practically challenging to solve the optimization.

\subsection{Network Architecture for Dimension Reduction} \label{sec:network arch}

To handle the permutation invariance and scalability required for our multi-agent setting, we employ the Deep Set architecture \citep{Zaheer2017DeepS} for both the policy and critic networks. This architecture ensures that the network is invariant to the order of input agents and can handle varying numbers of agents.
The permutation invariant network enjoys several advantages:
\begin{itemize}
    \item Permutation Invariance: The network's output remains unchanged regardless of the order of input agents.
    \item Permutation Invariance: The network's output remains unchanged regardless of the order of input agents.
    \item Input Size Irrelevance: The network can handle any number of agents without modification
    \item Privacy Consideration: Individual agent information is encoded into embeddings before being aggregated, ensuring privacy.
    \item Generalization: The trained network can be applied to new cohorts of agents with different sizes.
\end{itemize}

\begin{figure}[ht]
    \centering
    \includegraphics[width=0.7\linewidth]{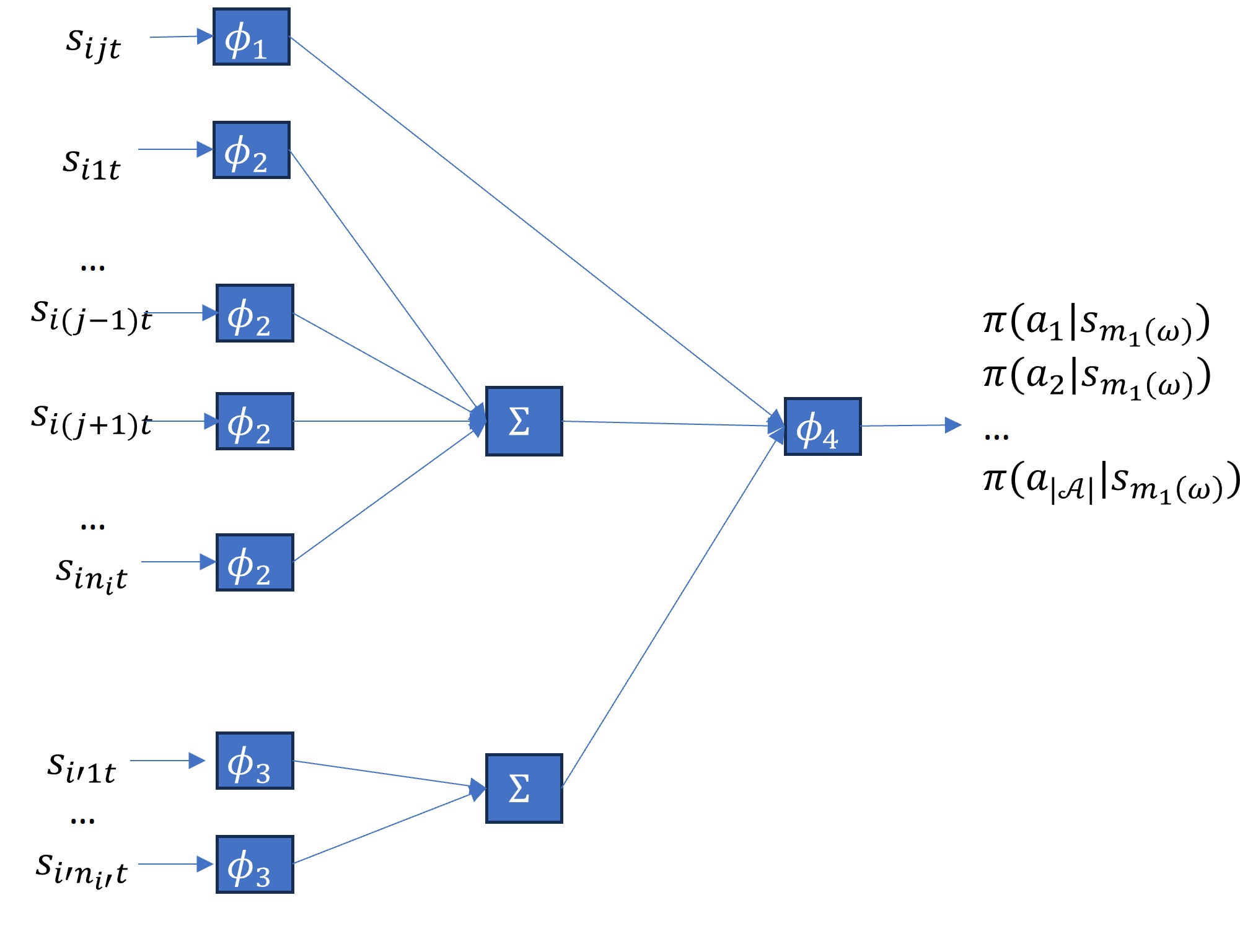}
    \caption{Permutation-Invariant Policy Network for subject $j$ from time $i$ at time $t$. Assume team $i$ is matched with team $i^\prime$.}
    \label{fig:PI network}
\end{figure}

Without loss of geneority, assume team $i=m_1(\omega)^1$ at time $t$. 
The decomposed individual policy $\pi(\aijt|s_{m_1(\omega),t}) = \pi(\aijt|\sijt, s_{m_1(\omega)^1}\setminus \sijt, s_{m_1(\omega)^2})$ is designed in this way. First the individual's state $\sijt$, the state of each of the rest individual from team $i$, and the state of each of the individuals from team $m_i(\omega)^2$ are separatly passed through three encoders. Then, the embeddings of the remaining individuals from team $i$ and team $m_1(\omega)^2$ are pooled together separately by some permutation invariant operators such as averaging to obtain the team member's characters and the rival characters. Finally, the individual embedding, the team member's characters and the rival characters are passed through some layers to output the final action for this individual. The policy network designed is illustrated in Figure \ref{fig:PI network}.

\begin{figure}[ht]
    \centering
    \includegraphics[width=\linewidth]{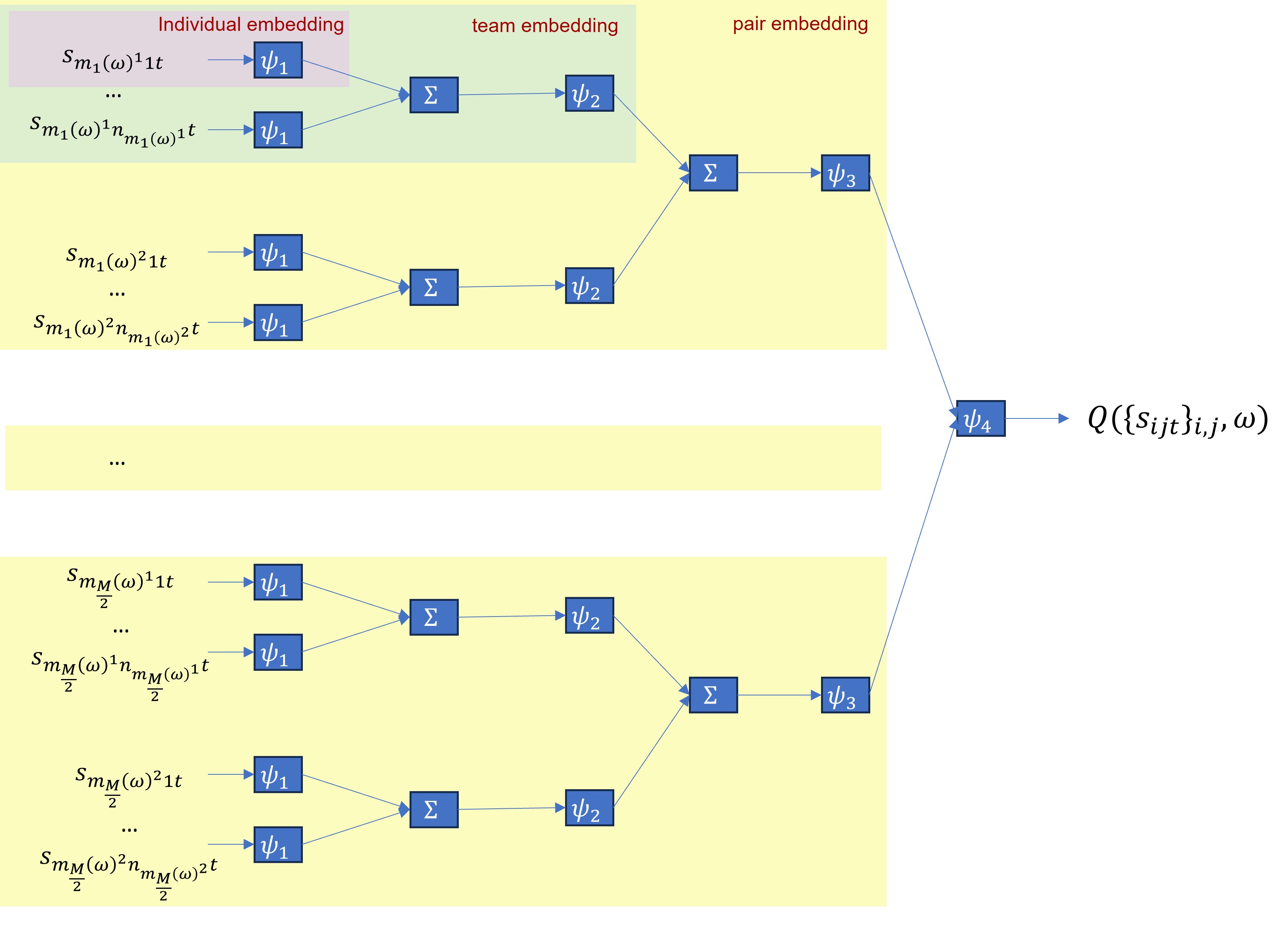}
    \caption{Permutation-Invariant Critic Network Design.}
    \label{fig:PI critic}
\end{figure}

The joint Q-network is designed to avoid explicit enumeration of the action space when combined with a greedy policy. It uses a shared encoder to generate embeddings for individual agents, which are then aggregated using permutation-invariant operators to form team and then group embeddings. This design significantly reduces the number of parameters and accelerates learning.
The joint Q-network is illustrated in Figure \ref{fig:PI critic}.
Given the Q-network architecture in Figure \ref{fig:PI critic}, the greedy policy over options can be derived by solving:
$$
\begin{aligned}
    & \max & \sum_{t_1=1}^M \sum_{t_2=1}^M a_{t_1,t_2} \psi_4(t_1, t_2) \\
    & \text{s.t.} & a_{t_1,t_2} \in \{0,1\} \\
    & & a_{t_1,t_2} = a_{t_2,t_1} \\
    & & \sum_{t_1=1, t_2 \neq t_1}^M a_{t_1,t_2} = 1 \quad \text{for } t_2 = 1, 2, \dots, M.
\end{aligned}
$$
where $\psi_4(t_1, t_2) = \psi_4\left(\psi_3\left(\psi_2\left(\sum_{i \in t_1} \psi_1(s_i)\right) + \psi_2\left(\sum_{i \in t_2} \psi_1(s_i)\right)\right)\right)$.

This formulation allows for efficient computation of the optimal groupings without explicitly enumerating all possible pairings.

\section{Simulation Study}
We design a simulated environment similar to the Intern Health Study.
The state variable for each individual is the daily the square root of step count and the reward is the square root of next day's step count. The action is a binary variable indicating whether to send a text message or not. We also allow some teams not enter competition in some weeks. We would view this case as a pair $(m_k(\omega)^1, m_k(\omega)^2)$ where $m_k(\omega)^1=m_k(\omega)^2$.
The transition function is defined as:
$$
\Sijt[m_k(\omega)^{q}][j][t+1] = \left\{
\begin{aligned}
    & -0.0005+ 0.3289\Sijt[m_k(\omega)^{q}] +  0.0672 \sum_{j^\prime} \Sijt[m_k(\omega)^{q^\prime}[j^\prime]] + 0.0103\mathbb{I}(q=q^\prime) \quad \text{if}\quad \Aijt = 0;\\
    & -0.0009+0.32450\Sijt[m_k(\omega)^{q}] +  0.0746 \sum_{j^\prime} \Sijt[m_k(\omega)^{q^\prime}[j^\prime]] + 0.0136\mathbb{I}(q=q^\prime) \quad \text{if}\quad \Aijt = 1;
\end{aligned}
\right.
$$
All encoders in the critic model and the policy network is a two-layer perceptron with each layer containing two nodes followed by a ReLu activation function.
We initialize 10 teams each containing 10 subjects with initial state sampled from a standard normal distribution. We train the policy with 10000 iterations and a learning rate of 0.001. After training, the obtained policy is evaluated in the same simulated environment by executing it for 1000 steps and calculated the cumulative reward. The gamma discounted factor is set ot be 0.9. We compare the policy obtained by the proposed method with two fixed policy (executing action 0 or 1) and a random policy which randomly match teams each week and choose action 0 or 1 for each individual with equal probabilities. The evaluated reward is reported in Table where the results are averaged over 20 repititions.
\begin{table}[htbp]
    \centering
    \begin{tabular}{ccc}
    \hline
    
    Method & Average reward & 0.9-discounted cumulated reward\\ \hline
        Porposed & -143.04 &-1436.97  \\
        Action 0 & -259.87 & 2601.24\\
        Action 1&-238.22 & -2391.28\\
        Random & -248.91 &-2495.50\\ \hline
    \end{tabular}
    \caption{The estimated value of different methods.}
    \label{tab:my_label}
\end{table}

\bibliographystyle{plainnat}
\bibliography{ref}

\end{document}

%% file: macro.tex
\newcommand{\ba}{\mathbf{a}}
\newcommand{\bs}{\mathbf{s}}

\NewDocumentCommand{\Rijt}{O{i}O{j}O{t}}{%
    R_{#1,#2,#3}
}
\NewDocumentCommand{\rijt}{O{i}O{j}O{t}}{%
    r_{#1,#2,#3}
}
\NewDocumentCommand{\fijt}{O{i}O{j}O{t}}{%
    f_{#1,#2,#3}
}
\NewDocumentCommand{\deltaijt}{O{i}O{j}O{t}}{%
    \delta_{#1,#2,#3}
}
\NewDocumentCommand{\xiijt}{O{i}O{j}O{t}}{%
    \xi_{#1,#2,#3}
}
\NewDocumentCommand{\epsilonijt}{O{i}O{j}O{t}}{%
    \epsilon_{#1,#2,#3}
}
\NewDocumentCommand{\Aijt}{O{i}O{j}O{t}}{%
    A_{#1,#2,#3}
}
\NewDocumentCommand{\sijt}{O{i}O{j}O{t}}{%
    s_{#1,#2,#3}
}
\NewDocumentCommand{\aijt}{O{i}O{j}O{t}}{%
    a_{#1,#2,#3}
}
\NewDocumentCommand{\Sijt}{O{i}O{j}O{t}}{%
    S_{#1,#2,#3}
}
\NewDocumentCommand{\TDijt}{O{i}O{j}O{t}}{%
    TD_{#1,#2,#3}
}

\NewDocumentCommand{\Ri}{O{i}}{%
    \mathbf{R}_{#1}
}
\NewDocumentCommand{\ri}{O{i}}{%
    \mathbf{r}_{#1}
}
\NewDocumentCommand{\Si}{O{i}}{%
    \mathbf{S}_{#1}
}
\NewDocumentCommand{\Ai}{O{i}}{%
    \mathbf{A}_{#1}
}
\NewDocumentCommand{\si}{O{i}}{%
    \mathbf{s}_{#1}
}
\NewDocumentCommand{\ai}{O{i}}{%
    \mathbf{a}_{#1}
}
\NewDocumentCommand{\myfi}{O{i}}{%
    \mathbf{f}_{#1}
}
\NewDocumentCommand{\TDi}{O{i}}{%
    \mathbf{TD}_{#1}
}
\NewDocumentCommand{\Yi}{O{i}}{%
    \mathbf{Y}_{#1}
}
\NewDocumentCommand{\Vi}{O{i}}{
    \mathbf{V}_{#1}
}
\NewDocumentCommand{\Di}{O{i}}{
    \mathbf{D}_{#1}
}
\NewDocumentCommand{\Gi}{O{i}}{%
    \mathbf{G}_{#1}
}
\NewDocumentCommand{\Omegai}{O{i}}{
    \mathbf{\Omega}_{#1}
}
\NewDocumentCommand{\Zi}{O{i}}{
    \mathbf{Z}_{#1}
}
\NewDocumentCommand{\gi}{O{i}}{%
    \mathbf{g}_{#1}
}
\NewDocumentCommand{\Qi}{O{i}}{%
    \mathbf{Q}_{#1}
}
\NewDocumentCommand{\Phii}{O{i}}{%
    \mathbf{\Phi}_{#1}
}
\NewDocumentCommand{\PhiLi}{O{i}}{%
    \mathbf{\Phi}_{L,#1}
}
\NewDocumentCommand{\psii}{O{i}}{%
    \mathbf{\psi}_{#1}
}
\NewDocumentCommand{\deltai}{O{i}}{%
    \mathbf{\delta}_{#1}
}
\NewDocumentCommand{\nSi}{O{i}}{%
    \mathbf{S'}_{#1}
}
\NewDocumentCommand{\conditionijt}{O{i}O{j}O{t}}{%
    S_{#1,#2,#3}, A_{#1,#2,#3}
}

\NewDocumentCommand{\tildePhiLijt}{O{i}O{j}O{t}}{%
    \widetilde{\Phi}_{L,#1,#2,#3}
}

\NewDocumentCommand{\tildegijt}{O{i}O{j}O{t}}{%
    \widetilde{g}_{#1,#2,#3}
}
\NewDocumentCommand{\tildeRijt}{O{i}O{j}O{t}}{%
    \widetilde{R}_{#1,#2,#3}
}

\NewDocumentCommand{\tildedeltaijt}{O{i}O{j}O{t}}{%
    \widetilde{\delta}_{#1,#2,#3}
}

\newcounter{assumptionCounter}
